\documentclass[3p,times]{elsarticle}

\usepackage{gensymb}

\usepackage{amssymb}
\usepackage{lineno,hyperref}
\usepackage{graphicx}
\usepackage{amsmath} 
\usepackage{color}
\usepackage{amsmath,mathtools}
\usepackage[table]{xcolor}
\usepackage{multirow}
\usepackage{booktabs}
\usepackage{subcaption}
\usepackage[numbers]{natbib} 
\usepackage{arydshln} 

\usepackage{mathtools}

\usepackage[figuresright]{rotating}

\begin{document}
	
	\bibliographystyle{elsarticle-num}

	\begin{frontmatter}
		
		\title{Automatic Surface Area and Volume Prediction on Ellipsoidal Ham using Deep Learning \tnoteref{mytitlenote}}
		
		
		\author[a]{Y.S. Gan}
		\author[b]{Sze-Teng Liong\corref{mycorrespondingauthor}}
		\cortext[mycorrespondingauthor]{Corresponding author. Department of Electronic Engineering, Feng Chia University, Taichung 40724, Taiwan R.O.C.}
		\ead{stliong@fcu.edu.tw}
		\author[c]{Yen-Chang Huang}
		\address[a]{Department of Mathematics, Xiamen University Malaysia, Jalan Sunsuria, \\Bandar Sunsuria, 43900 Sepang, Selangor, Malaysia}
		\address[b]{Department of Electronic Engineering, Feng Chia University, Taichung \\40724, Taiwan R.O.C.}
		\address[c]{School of Mathematics and Statistics, Xinyang Normal University, Henan, China}
		
		\begin{abstract}
			
			This paper presents novel methods to predict the surface area and volume of the ham through a camera.
			This implies that the conventional weight measurement to obtain the object's volume can be neglected and hence it is economically effective.
			Both of the measurements are obtained in following two ways: manually and automatically.
			The former is assumed as the true or exact measurement and the latter is through a computer vision technique together with some geometrical analysis that includes mathematical derived functions.
			For the automatic implementation, most of the existing approaches extract the features of the food material based on handcrafted features and to the best of our knowledge this is the first attempt to estimate the surface area and volume on ham with deep learning features.
			We address the estimation task with a Mask Region-based CNN (Mask R-CNN) approach, which well performs the ham detection and semantic segmentation from a video.
			The experimental results demonstrate that the algorithm proposed is robust as promising surface area and volume estimation are obtained for two angles of the ellipsoidal ham (i.e., horizontal and vertical positions).
			Specifically, in the vertical ham point of view, it achieves an overall accuracy of up to 95\% whereas the horizontal ham reaches 80\% of accuracy.   
		\end{abstract}
		
		\begin{keyword}
			\texttt{CNN,  Ham, Surface area, volume}
			\MSC[2010] 00-01\sep  99-00
		\end{keyword}
		
	\end{frontmatter}
	
	
	
	\section{INTRODUCTION}
	\label{sec:intro}
	
	The appearance of the a product is always one of the key factors for successful marketing in many occasions.
	Due to the emerging technologies and the rapid integration of software development, many online retailers opt to design attractive webpages in order to influence consumer perceptions of the web environment and thus leading to more sales and profits~\cite{prasad2009determinants}.
	An online 2017 UPS Pulse of the Online Shopper study with more than 6,400 European respondents from France, Germany, Italy, Poland, Spain, and the UK, discovered that 84\% of shoppers will shop in the physical store instead of online~\cite{ups2017}. 
	The major reason is that they can look around and touch and feel the products.
	In the in-store environment, McCabe and Nowlis~\cite{mccabe2003effect} found that the shoppers are less likely to pick up products with geometric properties as they highly rely on the sense of vision, which is believed to provide sufficient information of the products.
	For some of the products with geometric properties, especially packaged products, shoppers tends to evaluate the product through visually glancing instead of reach out or touch on the product~\cite{hoyer1984examination}.
	
	The geometry properties of an agricultural product is often indicated by its weight as it is relatively easy and quick to be measured with a digital scale.  
	An alternative solution to obtain the weight of the product is to measure its volume as the volume is a quantity defined by mass per unit density. 
	There are several works published to measure the volume of the objects by fusing the computer vision technique with some mathematical modeling and derivations.
	From the computer vision perspective, aside from benefiting the industrial automation performance, it is popularly utilized to tackle various real-world problems, such as biomedical image analysis~\cite{ronneberger2015u},  facial expression recognition~\cite{liong2018less}, gait tracking~\cite{pfister2014comparative}, phishing attack detection~\cite{rao2015computer}, and many other applications.
	Previous works applied computer vision to measure the volume of the food or agricultural product such as apple\cite{ziaratban2017modeling}, egg~\cite{soltani2015egg}, beans~\cite{anders2015numerical}, abalone~\cite{lee2015design} and ham~\cite{du2006estimating}.
	Note that many of the objects of interest are in ellipsoidal shape. 
	More analysis about the food measurement are discussed in~\cite{volumeoverview2015}. 

	Rasband~\cite{rasband2011imagej} proposes to utilize a three-dimensional scanner as a tool to capture the images of ellipsoidal bean and yellow lupine seeds. 
	Then, the author constructs the three-dimensional models~\cite{anders2015numerical} to obtain the volume of the object.
	The measurements are obtained from three distinct ways: micrometer, 3D scanner, and digital image analysis.
	Some of the geometric properties are first acquired by using the three methods, such as length, width, and thickness.
	Then, the values obtained are processed statistically at $\alpha$=0.05 and it shows that they are not significantly differed.
	However, when computing the surface area, there is a significantly difference for the surface area produced by the 3D scanner to that of micrometer and digital image analysis measurement.
	
	Lee et al.~\cite{lee2015design} design a laboratory-scale system to automatic measure several geometric parameters (i.e., total length, body length and thickness) of an abalone.
	The sample size in the experiment is 500 and they are ranged around 15$g$ to 130$g$.
	The 2D images are obtained from two CCD camera which installed at the base and the sides of the abalone. 
	The values are then derived to compute the volume of the abalone using a regression analysis.
	Concretely, in the computer vision automated methodology, an edge detection technique~\cite{davis1975survey} is applied on each image. 
	In order to determine the possible ellipse, a least square method is adopted~\cite{rosin1993note} to fit 100 of the edge points detected.
	As a result, the correlation coefficient of the calculated volume and the actual volume is 0.998.
	Besides, they demonstrated that using a regression analysis, the slope of the linear curve regression for the calculated volume against the actual volume is close to 1.
	
	In the recent years, artificial neural network (ANN) are applied into the machine vision system to estimate the weight of the food material.
	Asadi and Roufat~\cite{asadi2010egg} design a computer vision system with ANN methodology to predict the weight of an egg.
	For each egg sample, two images are captured and there are a total of twelve features extracted from each image to further enhance the features by adopting a Multi-Layer Perceptron (MLP) network in order to determine the weight of the egg.
	Their proposed method generates the correlation coefficient between the predicted and the actual weight of 0.96 and a maximum absolute error of 2.3$g$.
	
	A few years later, an improved machine vision system is proposed by Soltani et al.~\cite{soltani2015egg} to predict the volume in two different methods: the mathematical model based on Pappus' theorem~\cite{thomas2005thomas} and an ANN model.
	For the mathematical model, the correlation coefficient and the maximum absolute error values are 0.99 and 1.69 $cm^3$, respectively.
	The correlation coefficient obtained using ANN is 0.99 and the maximum absolute error is 2.06 $cm^3$. 
	Therefore, it demonstrates that the method proposed is better than that of~\cite{asadi2010egg} approach.
	The ANN consists of 28 neurons in the single hidden layer and input data to the ANN architecture is the major and minor diameter of the egg.

	On the other hand, an interesting mobile application is developed by Waranusast et al.~\cite{waranusast2016egg} to estimate and classify the egg size into six categories (i.e., pewee, small, medium, large, extra large and jumbo sizes) which is determined by their weight (i.e., varies from 45$g$ to more than 70$g$).
	The feature extractors utilized in the proposed algorithm are handcrafted representation-based approach, such as the Histogram of Oriented Gradients (HOG)~\cite{dalal2005histograms} object detector, GrabCut~\cite{rother2004grabcut} segmentation technique and Otsu thresholding~\cite{otsu1979threshold} method.
	By simply capturing the photo of the egg with a specific type of coin as the reference object, the egg size category can be acquired.
	In average, 80.4\% of classification accuracy is achieved. 
	
	To the best of our knowledge, there is only one attempt in the literature estimates the 
	volume of a ham that adopts a image processing algorithm~\cite{du2006estimating}.
	The authors assume the ham is having the shape of the ellipsoid.
	In order to achieve the objective, three measurements are first obtained from the image, viz., length, width and thickness of the ham.
	There are nine ham samples in the experiment and four images of each sample are captured with \degree{90} rotation increments around the major axis.
	Therefore, there are a total of four images taken for each ham sample.
	Then, the surface area and volume are estimated based on both the two mathematical derivations: partitioned method and derived method. 
	The object's shape feature extractors utilized in~\cite{du2006estimating} are well-known handcrafted descriptors.
	However, there are several essential parameters required to be heuristically tuned in order to optimize the prediction performance.
	For instance, the median filter size of the shape descriptor, Canny edge detector~\cite{canny1986computational} and threshold value of the edge detector, cubic spline interpolation technique~\cite{de1978practical}.
	After fine-tuning to the best settings, the highest accuracy for volume prediction is up to 98.21\% for the partitioned method.
	It should also be noted that the volume of the nine ham samples are quite similar, which has a range of approximately $920cm^3$ to $1060cm^3$, with the mean of $\sim 920cm^3$.
	
	In short, there are two major approaches to extract the meaningful features from input data, i.e. the traditional handcrafted feature extraction and automatic feature extraction (employing the deep learning architectures). 
	In contrast to the method proposed in~\cite{du2006estimating}, this paper proposes a novel automatic approach to estimate the volumes of hams using a deep learning model.
	There are three main advantages of exploiting the deep learning features rather than the handcrafted features:
	a) The characteristics of the input images are learned intuitively from the image data and thus consists both the low-level and high-level features;
	b) The feature representation has better generalization ability and is applicable on different data distributions, and;
	c) The need for feature engineering is highly reduced which is the most time-consuming parts in the machine learning practice.
	
\begin{figure}[t!]
		\centering
		\includegraphics[width=1\linewidth]{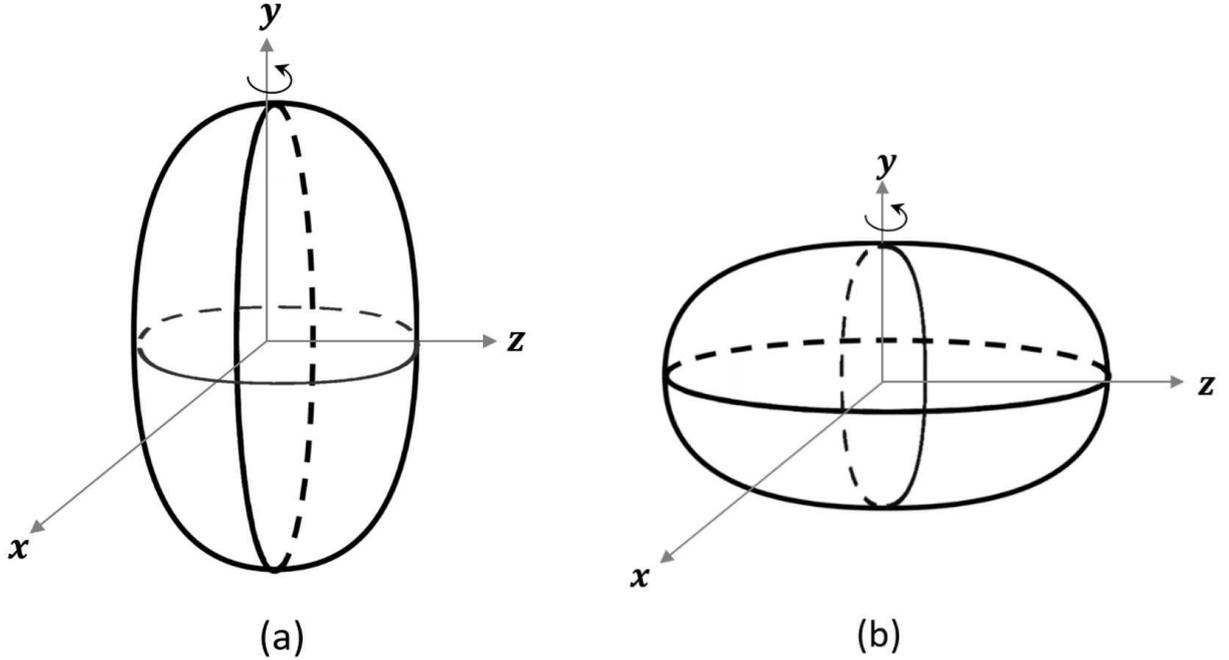}
		\caption{Schematic representations of the observatory viewpoints of (a) vertical and (b) horizontal.}
		\label{fig:xyz}
	\end{figure}

	The four contributions of this paper are briefly listed as follows:
	\begin{enumerate}
		\item Proposal of a feature extraction with deep learning technique that exploits the Mask-RCNN (Regional Convolutional Neural Network) instance segmentation model.
		The model is constructed and customized by training the hyper parameters from the ham samples.
		
		\item Both the vertical and horizontal viewpoints of the same ham are captured by the camera to verify the robustness of the proposed methodology. 
		
		\item Two numerical methods are derived to further analyze and estimate the volumes of hams: the vertical and horizontal approaches, as illustrated in Figure~\ref{fig:xyz}.
		Instead of assuming the hams being ellipsoid-shaped (for instances~\cite{du2006estimating, OKT}), the following assumptions are generalized, that depending on the observatory viewpoints:
		\begin{enumerate}
			\item Vertical - the hams are assumed to be rotationally symmetric about a fixed rotation axis.
			\item Horizontal - the proposed method can be applied on the non-rotationally symmetric objects (the mathematical derivation is explained in detail in Section \ref{sec:mathderiva}). 
		\end{enumerate}
		
		\item Three volume estimations of each ham are presented and statistically compared. Thorough analysis and discussion are provided to highlight the performance results.
	\end{enumerate}

	By comparing the existing methods in estimating the volumes of geometric objects, the mathematical formula (i.e., Equation~ \ref{eqn4_5}) derived in Section \ref{sec:mathderiva} is an appropriate application of Integral Geometry to rotational objects. 
	In contrast to most of the works~\cite{anovelmethodegg, MonteCarlo1, MonteCarlo2} that obtain the volume of the food material by employing the Monte Carlo method of random points of boundaries, we integrate the surface of revolutions and partition method to perform the volume estimation.
	Moreover, the volume formula (3) in~\cite{OKT} or~\cite{deterorange1} appeared to be more adequate methods that can apply on sphere-like symmetric objects.
	In addition, the two cameras are positioned such that it has a view angle directed perpendicularly at the object.
	However, such a set up restricts the flexibility of the position of cameras.
	On the contrary, the derived formula (i.e., Equation~ \ref{eqn4_5}) seems to be an optimal solution for volume estimation in this study.
	Concretely, it needs only a single camera and a sample with multiple angles of view (i.e., by rotating the sample).
	Hence, it is highly cost effective and can be implemented more efficiently.
	As a side note, we are aware that Pappus' theorem~\cite{schwartz1993pappus} (that also utilized in \cite{soltani2015egg}) can be applicable to our experiments (i.e. the vertical case), except for the horizontal ham.

	The remainder of the paper is organized as follows.
	Section~\ref{sec:procedure} describes the procedure of developing the ham database, including the materials and methods of eliciting the ham videos sequences. 
	Section~\ref{sec:CVS} develops the computer vision system and how we use deep learning for detection.
	In addition, the framework of the proposed algorithm is introduced and elaborated. 
	In Section \ref{sec:mathderiva}, the mathematical derivations for the volume of hams are presented. 
	Note that in the derivation, we do not assume that the hams are ellipsoid-shaped. 
	Section~\ref{sec:result} reports and discusses the experimental results.
	Lastly, Section~\ref{sec:conclusion} concludes the paper.
	
	\section{Elicitation Procedure}
	\label{sec:procedure}
	\subsection{Experiment Protocols}
	\label{subsec:protocol}
	
\begin{figure}[t!]
		\centering
		\includegraphics[width=1\linewidth]{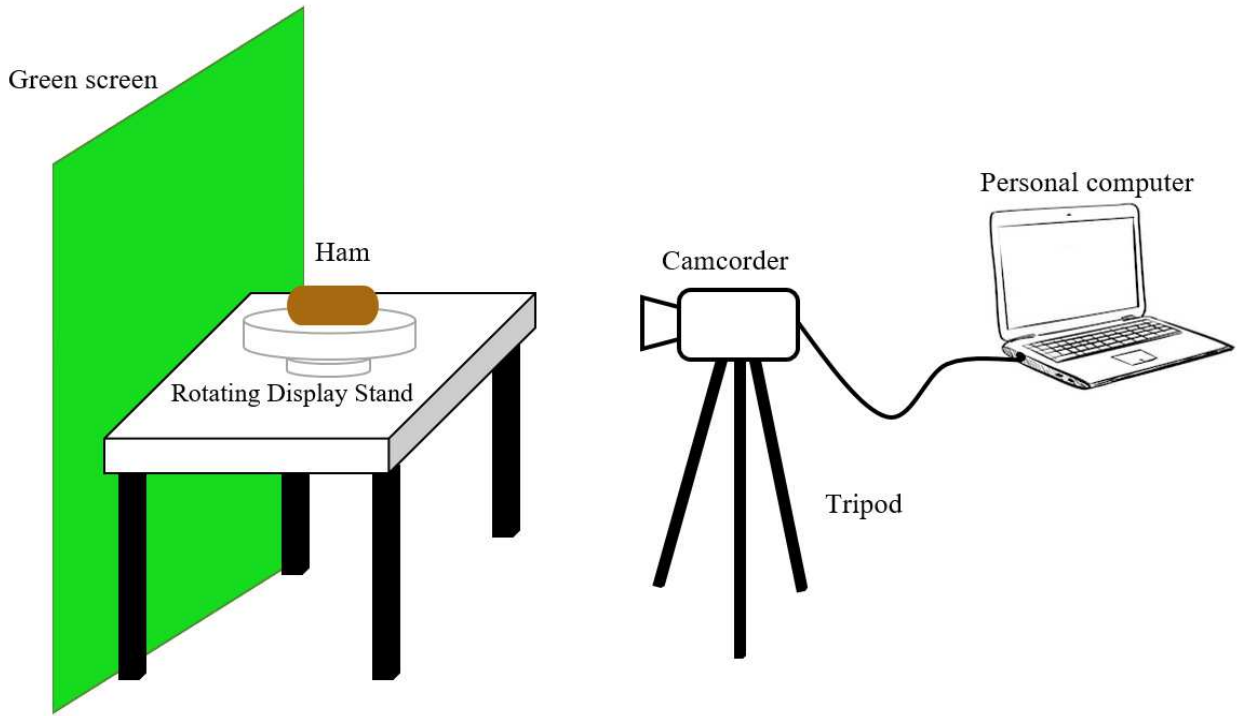}
		\caption{Acquisition setup for elicitation and recording of the ham.}
		\label{fig:setup}
	\end{figure}

	The experiment comprised of four types of hams and we select two ham samples for each of the same type.
	Therefore, there are a total of eight hams.
	The apparatus consists of of camcorder, tripod, rotating display stand, green screen and a personal computer, as depicted in Figure~\ref{fig:setup}.
	The videos are recorded using Sony FDR-AX100 and the details of the camcorder are described in Table~\ref{table:sony}.
	The angle and the height of the camcorder with a tripod stand adjusted such that it is perfectly parallel aligned to the the ham's surface, where the distance from the camcorder to the sample is set to 36.8cm.
	The camcorder is connected to a personal computer via a USB cable so that the videos captured from the camcorder can be automatically transfer and save into the computer in a real-time recording.
	A green backdrop is used as the background as it is the common practice in most of the film shooting.
	This is due to the background is easily to be digitally modified using a technique called chroma keying, such as to remove the background by rendering it transparent.
	A rotating display stand is placed on a table to provide 360 degree display accessibility of the ham to the camcorder.
	In addition, the rotating display stand is turning consistently with constant speed values (i.e., 15$s$/ circle or 20$s$/ circle) as it is electricity (i.e., battery) operated.
	We set the speed of the rotating display stand to 20$s$/ circle to allow more angles of the ham to be captured by the camcorder and a it is set to revolve in clockwise direction.
	
\begin{figure*}[t!]
		\centering
		\includegraphics[width=1\linewidth]{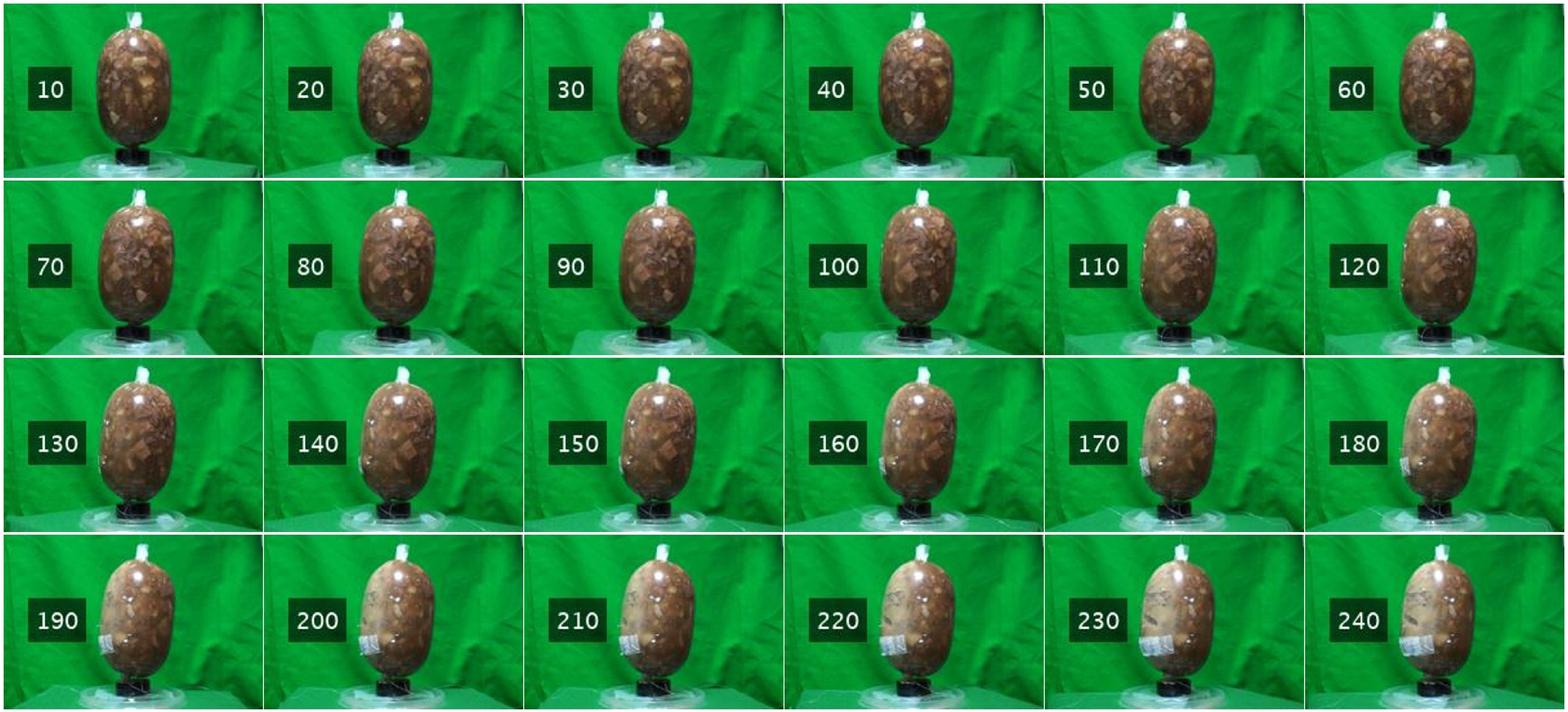}
		\caption{A demonstration of the frame sequence of the ham when it is placed horizontally on the rotating display stand. The numeric value beside the ham indicates the frame number.}
		\label{fig:2_1}
	\end{figure*}

\begin{figure*}[t!]
		\centering
		\includegraphics[width=1\linewidth]{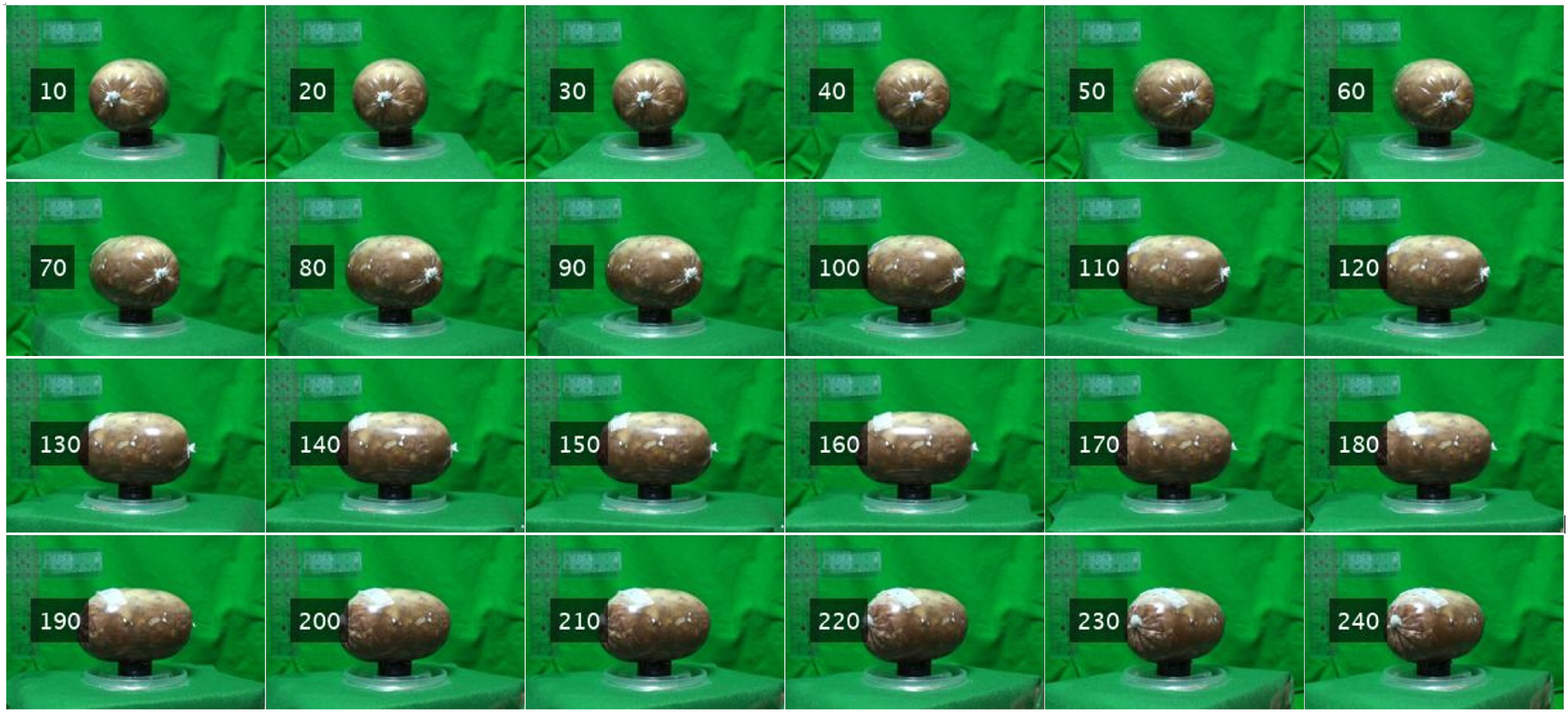}
		\caption{A demonstration of the frame sequence of ham horizontally on the rotating display stand. The numeric value beside the ham indicates the frame number. }
		\label{fig:2_2}
	\end{figure*}

	Figure~\ref{fig:2_1} and Figure~\ref{fig:2_2} illustrate parts of the frame sequence when eliciting the ham vertically and horizontally, respectively.
	Note that, in order to increase the contrast of the foreground (i.e., ham) and the background, we try to cover all the apparatus with green clothes.
	This is aim to improving the distinguish ability of the ham object especially when applying the automatic object detection system.
	Besides, the average duration of each video is 60$s$, which means the ham is rotated for at least three times.
	Therefore, the number of images can be extracted is more than 1800 frames.
	This is to allow us to construct to a better model architecture with more training data.
	
	\setlength{\tabcolsep}{5pt}
	\begin{table}[h]
		\begin{center}
			\caption{Camcorder specifications and configurations}
			\label{table:sony}
			\begin{tabular}{lc}
				\noalign{\smallskip}
				\hline
				\noalign{\smallskip}
				Feature
				& Description \\
				\hline
				
				\noalign{\smallskip}
				Resolution (pixels)
				& 1920 $\times$ 1080 \\
				
				\noalign{\smallskip}
				Frame rate (fps)
				& 30\\
				
				\noalign{\smallskip}
				Shutter Speed (s)
				& 1/60\\
				
				\noalign{\smallskip}
				Sensor Type
				& 1-Chip 1/2.5" CMOS\\
				
				\noalign{\smallskip}
				Focus Adjustment
				& Automatic\\
				
				\noalign{\smallskip}
				Light Exposure Mode
				& Automatic\\
				
				\hline
				
			\end{tabular}
		\end{center}
	\end{table}

	\section{Computer Vision System}
	\label{sec:CVS}
	
	Mask-RCNN~\cite{he2017mask} is a deep neural network which can efficiently identify the bounding box of an object as well as segment the object instance concurrently. 
	It is an extension of Faster R-CNN~\cite{ren2015faster}, a popular object detection framework.
	A parallel branch for predicting an object mask (i.e., the segmentation boundary of the object) is added to the original Faster R-CNN model.
	Owing to the advantages of the Mask-RCNN algorithm that are conceptually simple, high speed, high instance segmentation accuracy and easily to be implemented, it has been widely adopted in several applications.
	For instance, multi-person pose estimation~\cite{girdhar2018detect}, disruptions detection to the patient’s sleep-wake cycle~\cite{malhotra2018autonomous},  fine-grained image recognition~\cite{wei2016mask},  localizing nuclei in medical imaging~\cite{loudon2018detecting}, etc.
	
	In Section~\ref{subsec: background}, we discuss the details of Mask-RCNN architecture as well as its theoretical implications.
	Then, Section~\ref{subsec:implement} elaborates the implementation of Mask-RCNN into our experiment.
	Specifically, the neural network mechanism consists four main steps: mask annotation, hyper parameter settings initialization, model training and model testing. 
	
	\subsection{Background of Mask-RCNN}
	\label{subsec: background}
	Concretely, the Mask-RCNN architecture is constructed based on a Feature Pyramid Network (FPN)~\cite{lin2017feature} and a ResNet-101~\cite{he2016deep} backbone.
	FPN simply takes in a single-scale input.
	The architecture is designed such that it is a top-down architecture with multiple lateral connections.
	By capturing and integrating the image's information from the feature pyramid at different layers, it enables to generate a richer semantic map to represent the input image.
	On the other hand, ResNet-101 is a convolutional neural network (CNN) that initially was trained on ImageNet database~\cite{imagenet_cvpr09}, where this database composes more than a million images with 1000 object categories.
	However, noted that the pre-trained model of ResNet-101 is designed for object recognition task, the instance segmentation task is not considered when training the architecture.
	Therefore, the parameters (i.e., weights and biases) of the Mask-RCNN are then retrained using the Microsoft Common Objects in Context dataset (MS COCO)~\cite{lin2014microsoft}.
	In brief, this dataset contains up to 1.5 million labeled object instances with 80 object categories.
	
	The overall methodology in Mask-RCNN is listed as follows:
	\begin{enumerate}
		\item  The input image is fed into the CNN  for feature extraction;
		\item Several feature maps with different spatial resolutions are generated by FPN;  
		\item  The regions of interest (RoIs) to fixed size input are produced by utilizing the RoI alignment (RoIAlign), in which it approximates the features via bilinear interpolation for better object localization.
		\item A three parallel tasks are performed by utilizing the RoIAligned features, where the tasks include the bounding box regression, classification and mask prediction.
		
	\end{enumerate}

	\subsection{Mask-RCNN implementation}
	\label{subsec:implement}
	The method proposed in this paper is modified based on the Mask-RCNN architecture design. 
	The implementation is done using Tensorflow and ResNet101 as the backbone network. 
	However, the ResNet101 architecture configuration is trained from scratch and the parameters are fine-tuned to adapt into our experimental scenario.
	The retrain procedure is a necessity and solely performing the transfer learning may not be a good choice in our case due to the two main reasons: 
	1) The object category of the pre-trained model of architecture does not contain ham. 
	Therefore, it is reasonable that the instance segmentation of the target ham will be inaccurate and imprecise.
	2) Since the dataset in this experiment is relatively small in size (i.e, total of 8 hams), it is most likely to trigger the overfitting phenomena.

	To construct a robust instance segmentation architecture that can precisely detect the ham location and provide its boundary's coordinate, we carry out the following four major steps:

	\begin{figure*}[h]
		\centering
		\includegraphics[width=1\linewidth]{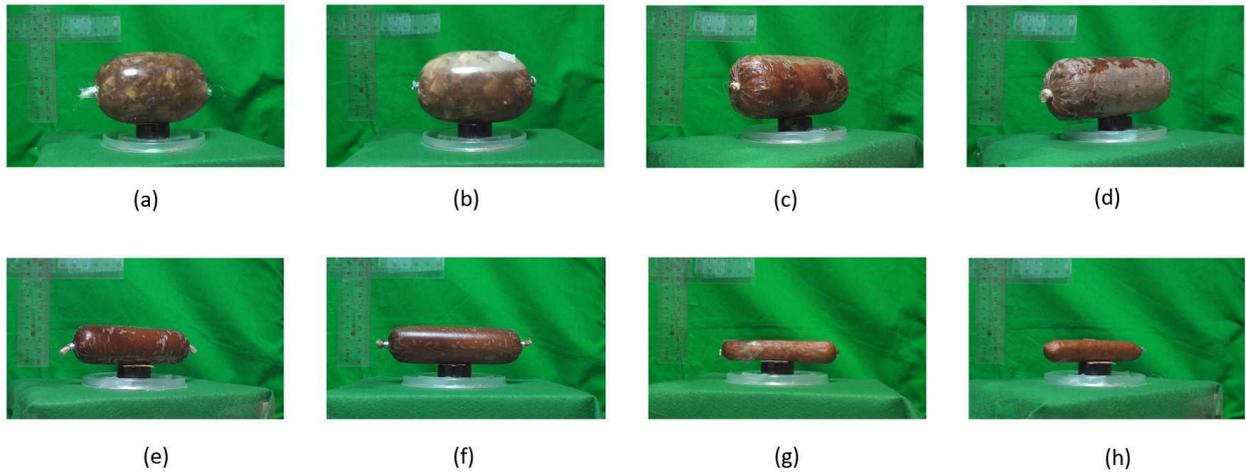}
		\caption{Eight hams for validation experiments, where the tag label are: (a) ham 1, (b) ham 2, (c) ham 3, (d) ham 4, (e) ham 5, (f) ham 6, (g) ham 7, (h) ham 8.}
		\label{8ham}
	\end{figure*}
	
	\begin{figure*}[h]
		\centering
		\includegraphics[width=0.5\linewidth]{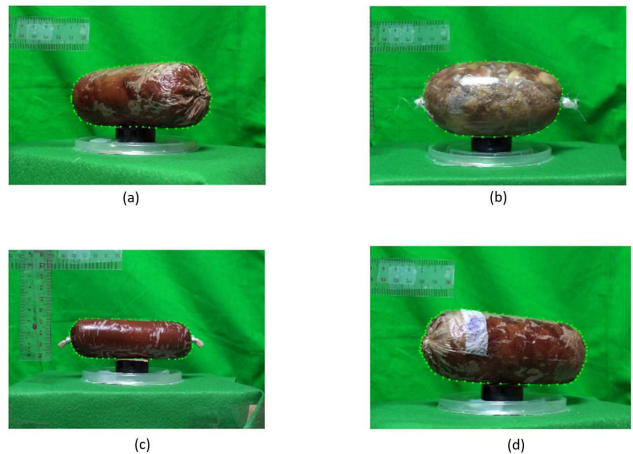}
		\caption{Manual annotations on the ham samples, where the annotated item is enclosed by the green dots.}
		\label{dotted}
	\end{figure*}
	
	\begin{figure*}[h]
		\centering
		\includegraphics[width=0.5\linewidth]{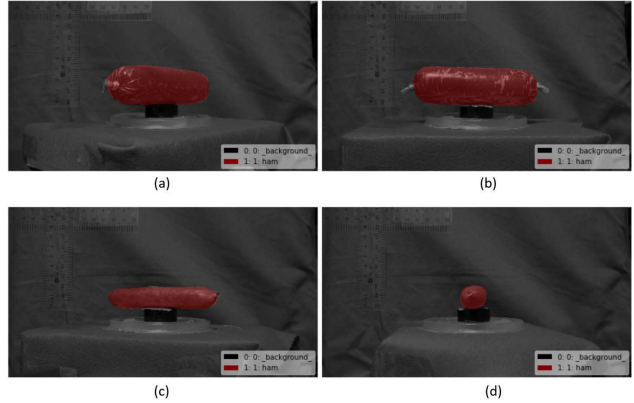}
		\caption{The instance segmentation results in the architecture training process. The region with red color is the detected object or the foreground, while the black region denoting the background. }
		\label{masked_trained}
	\end{figure*}

	\begin{enumerate}
		\item Mask Annotation \\
		As mentioned in Section~\ref{subsec:protocol},  the dataset in this experiment consists of 8 hams, as shown in Figure \ref{8ham}. 
		To increase the size of training samples, the images of the ham are collected at different angles.
		The ham is placed a rotating display stand for recording while the base rotates staystill. 
		Then, a manual ground-truth annotation of the region of ham is performed on the ham images.
		The annotation tool utilized in the experiment is~\cite{matterport_labelme_2016}, which enables us to define the irregular regions of the target object in an image.
		The sample annotated binary masked image is illustrated in Figure~\ref{dotted} with reasonably precise pixel-wise closed boundary.
		Note that a high quality annotation with more precise fine edges object outline can be achieved with this tool by clicking more interest points around the object.
		By doing this, it is most likely to lead to high prediction accuracy but it is very time-consuming.\\
		
		\item Hyper Parameters Settings Initialization\\
		The Mask-RCNN model is implemented on Python 3.6.
		The two specific Python libraries exploited for the feature learning task are Keras and TensorFlow. 
		Specifically, we leverage the Matterport Github repository~\cite{matterport_maskrcnn_2017} to build custom model.  
		The reasons being are: it is one of the most popular frameworks, easy to use and very well documented.
		This Mask-RCNN model is based on Feature Pyramid Network and ResNet101 backbone. 
		The repository provided includes the: 
		a) Source code of Mask-RCNN built on FPN and ResNet101;
		b) Training code for MS COCO;
		c) Pre-trained weights for MS COCO;
		d) Jupyter notebooks to visualize the detection pipeline at every step, and;
		e) Parallel Model class for multi-GPU training. 
		Before training the model, the MS COCO pre-trained model and the pre-trained weights are first adopted.
		Then, the number of object classes is updated, whilst keeping most of the model parameters the same.

		The details of the architecture and setting values are listed in Table~\ref{table:configuration}.
		\\

		\setlength{\tabcolsep}{5pt}
		\begin{table}[h]
			\begin{center}
				\caption{Configuration type and Parameters of Mask-RCNN model}
				\label{table:configuration}
				\begin{tabular}{lc}
					\noalign{\smallskip}
					\hline
					\noalign{\smallskip} Configuration type & Parameter \\
					\hline
					
					\noalign{\smallskip} Backbone strides    & $[4, 8, 16, 32, 64]$ \\
					\noalign{\smallskip} Batch size   & 1 \\
					\noalign{\smallskip} bbox standard deviation     &  $[0.1 0.1 0.2 0.2]$\\
					\noalign{\smallskip} Compute backbone shape  &  None \\
					\noalign{\smallskip} Detection Max Instances      &    100\\
					\noalign{\smallskip} Detection Min Confidence     &    0.7\\
					\noalign{\smallskip} Detection Nms Threshold      &    0.3\\
					\noalign{\smallskip} Fpn Classif Fc Layers Size   &    1024\\
					\noalign{\smallskip} Gpu Count                  &      1\\
					\noalign{\smallskip} Gradient Clip Norm          &     5.0\\
					\noalign{\smallskip} Images Per Gpu             &      1\\
					\noalign{\smallskip} Image Max Dim              &      1024\\
					\noalign{\smallskip} Image Meta Size            &      14\\
					\noalign{\smallskip} Image Min Dim              &      512\\
					\noalign{\smallskip} Image Min Scale           &       0\\
					\noalign{\smallskip} Image Resize Mode          &      square\\
					\noalign{\smallskip} Image Shape                 &     $[1024 1024    3]$\\
					\noalign{\smallskip} Learning Momentum          &      0.9\\
					\noalign{\smallskip} Learning Rate             &       0.001\\
					\noalign{\smallskip} Mask Pool Size             &      14\\
					\noalign{\smallskip} Mask Shape                 &      [28, 28]\\
					\noalign{\smallskip} Max Gt Instances           &      100\\
					\noalign{\smallskip} Mean Pixel                 &      [123.7 116.8 103.9]\\
					\noalign{\smallskip} Mini Mask Shape             &     (56, 56)\\
					\noalign{\smallskip} Name                       &      shapes\\
					\noalign{\smallskip} Num Classes                 &     2\\
					\noalign{\smallskip} Pool Size                  &      7\\
					\noalign{\smallskip} Post Nms RoIs Inference    &      1000\\
					\noalign{\smallskip} Post Nms RoIs Training      &     2000\\
					\noalign{\smallskip} RoI Positive Ratio         &      0.33\\
					\noalign{\smallskip} RPN Anchor Ratios         &       $[0.5, 1, 2]$\\
					\noalign{\smallskip} RPN Anchor Scales         &       $(96, 192, 384, 768, 1536)$\\
					\noalign{\smallskip} RPN Anchor Stride         &       1\\
					\noalign{\smallskip} RPN bbox Std Dev           &      $[0.1 0.1 0.2 0.2]$\\
					\noalign{\smallskip} RPN NMS Threshold          &      0.7\\
					\noalign{\smallskip} RPN Train Anchors Per Image  &    256\\
					\noalign{\smallskip} Steps Per Epoch              &    100\\
					\noalign{\smallskip} Top Down Pyramid Size       &     256\\
					\noalign{\smallskip} Train BN                  &       False\\
					\noalign{\smallskip} Train RoIs Per Image      &       33\\
					\noalign{\smallskip} Use Mini Mask             &       True\\
					\noalign{\smallskip} Use RPN RoIs              &       True\\
					\noalign{\smallskip} Validation Steps         &        10\\
					\noalign{\smallskip} Weight Decay              &       0.0001\\

					\hline
				\end{tabular}
			\end{center}
		\end{table}

		\item Model Training\\
		To train the model, MS COCO pre-trained model is used as the checkpoint to perform transfer learning.
		A total number of 560 sample images of ham with different rotation angles are selected as the training set.
		The Matterport~\cite{matterport_maskrcnn_2017} algorithm is modified by adding the ham images and annotation into a new object category.
		The model is trained on a GPU for about 60 epochs and the training process is completed in three and a half hours.  
		It is worth highlighting that the neural network training process will be terminated when the training loss and validations loss was zig-zagging. 
		Note that all models were trained in an end-to-end fashion.
		In addition, both the horizontally and vertically rotating positions are taking into account to build the training model.\\
		
		\item Model Testing\\
		All the three-minutes videos of the eight hams with both horizontal and vertical positions are treated as the testing datasets.
		The images are first extracted from each of the video and will be fed into the architecture we trained in the aforementioned step.
		The trained Mask-RCNN is expected to predict the location of the instance by outlining the object's boundary.
		The sample output of a testing image is shown in Figure~\ref{masked_trained}.
	\end{enumerate}

	All the experiments were carried out on an Intel Core i7-8700K CPU 3.70 GHz processor with 
	The experiments  were  carried out by Python 3.6 in  Intel Core i7-8700K 3.70 GHz processor, RAM  32.0 GB,  GPU NVIDIA GeForce GTX 1080 Ti.
	
	\section{Mathematical Derivations}
	\label{sec:mathderiva}
	This section describes the details on the mathematical foundation, derivation and analysis in deriving the equations for the volume of the ham, in both vertical and horizontal positions.
	Consider the ham is having the shape of ellipsoid in the orthogonal $xyz$-coordinates, whose origin is located at the center of the ellipsoid.
	In the three-dimensional Euclidean space $R^3$, we deploy the ellipsoids in two different positions: vertical and horizontal, as illustrated in the schematic diagrams in Figure~\ref{fig:xyz} (a) and (b) respectively.
	However, it is generally acknowledged that the water displacement method (i.e., the Archimedes principle), is still the best way to measure the volume of an irregular object.
	Notice that the water displacement method is employed in this experiment to serve as the reference value..
	This is also to evaluate the correctness of the proposed method. 
	The procedure to perform the manual measurement is elaborated in Section~\ref{subsec: water}. 
	In spite of that, such procedure is inconvenient, difficult, or sometimes impossible to perform in many of the circumstances due to the practical limitations.
	For example, in the scenarios of the automated production lines or the when tested on non-waterproof for experimental objects. 
	
	Instead of using the measurement method, there are plenty of formulas for calculating the surface area and volume of an object.
	For example, one of the popular methods to compute the surface area is known as the Knud Thomsen's approximation (\cite{kresta2015advances}, page 524).
	Unfortunately, it is not applicable in our case as it requires the lengths of the semi-axes of the object.
	As such, we derive the surface area and volume calculations that the prior knowledge of the semi-axes for the geometry estimations is not needed.
	Concisely, a numerical solution is proposed, which is obtained through the revolution of arbitrary bi-dimensional geometries of the ham.
	Particularly, the specific revolution of the two-dimensional plane surfaces about a fixed axis in the same plane are the vertical and horizontal angle, that represent the parallel rotational to the y-axis and x-axis, respectively.
	The detail of the mathematical derivations and descriptions of both the dimensions to calculate the surface area and volume are described in two subsections below.

	Now we introduce and describe the notations which are used in the subsequent sections. 
	There are eight samples (i.e., hams) in total and the video clip of the $j$-th ham ($j=1,\cdots, 8$) is denoted by:
	
	\begin{equation}
	\mathbf{U}^j= \{f_k| k=1,\cdots, N_j\}
	\end{equation}
	where $N_j$ the total number of frames in the video of the $j$-th ham. 
	Since the following steps are appropriate to describe for all hams, 
	the superscript index $j$ of $U^j$ and the subscript $j$ of $N_j$ are sometimes ignored. 
	
	\subsection{The long axis parallel to the $y$-axis (vertical case, Figure \ref{fig:xyz} (a).)}
	Given a function $f(y)$ defined on the $xy$-plane for $y \in \left[a, b \right]$. In Calculus, it is known that the surface area $S(\Sigma)$ and the volume $V(\Sigma)$ for the solid of revolution $\Sigma$ by rotating the function $f(y)$ about the $y$-axis can be represented as:
	
	\begin{equation}\label{eqn4_1}
	S(\Sigma) = \int_{a}^{b}{2 \pi f(y) \sqrt{1+[f'(y)]^2} dy}
	\end{equation}
	
	\noindent and
	
	\begin{equation}\label{eqn4_2}
	V(\Sigma) = \int_{a}^{b}{2 \pi [f(y)]^2 dy}.
	\end{equation}
	
	In our experiment, suppose the video of the ham consists of $N$ number of frames. In each frame of the ham's video, we calculate the surface area and the volume of the solid of revolution by rotating
	the boundary of the ham and using Equations \ref{eqn4_1} and \ref{eqn4_2}. 
	Finally, the surface area and the volume of the ham are obtained by taking the average sum of the $N$ number surface areas and volumes of the solid of revolutions in all frames. 
	
\begin{figure}[t!]
		\centering
		\includegraphics[width=0.4\linewidth]{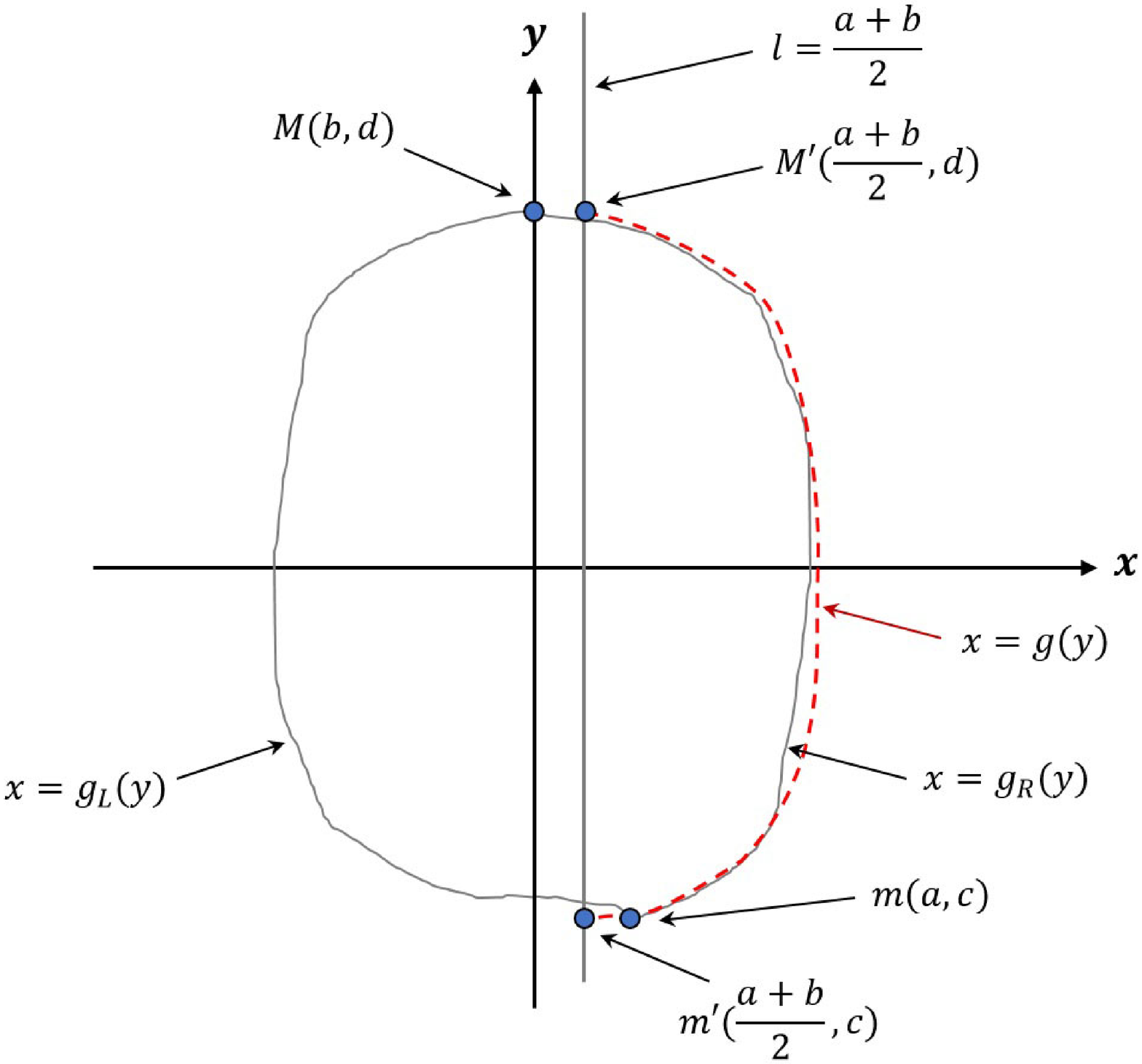}
		\caption{An illustrated shape of convex ham for mathematical modeling.}
		\label{fig:convex}
	\end{figure}

	Given a nearly convex domain $\Sigma$ with boundary $\partial \Sigma$ on the $i$-th picture (as shown in Figure \ref{fig:convex}), let $m(a,c)$ and $M(b,d)$ be the points on the boundary $\partial \Sigma$ with minimal and maximal $y$-coordinates respectively.
	Let the equations of $m'$ and $M'$ to be:
	\begin{equation}\label{eq:m}
	m' = m'(\frac{a+b}{2}, c)
	\end{equation}
	
	\noindent and
	\begin{equation}\label{eq:M}
	M' = (\frac{a+b}{2}, d).
	\end{equation}
	
	\noindent Consider the vertical line as the rotation axis, and it can be expressed as:
	
	\begin{equation}\label{eq:l}
	l: x = \frac{a+b}{2}.
	\end{equation}

	\noindent The line $l$ is therefore separating the boundary $\partial \Sigma$ into two parts, denoted by the left curve $x = g_L(y)$ and the right curve $x = g_R(y)$. 
	It is noted that the two curves only meet at the points $m'$ and $M'$ and $g_R(y) \ge g_L(y)$ for any $y \in [c, d]$. 
	Next, let the following equation as the boundary function on the right-hand-side of $l$:
	
	\begin{equation}\label{eq:l}
	g(y): = \frac{a+b}{2} + \frac{g_R(y) - g_L(y)}{2}.
	\end{equation}
	
	\noindent The curve $x = g(y)$ (viz., the dashed line in Figure~\ref{fig:convex}) is the average horizontal distances of $g_R(x)$ and $g_L(x)$ to the rotation axis $l$. 
	Thus, by substituting $x = g(y)$ into Equations \ref{eqn4_1} and \ref{eqn4_2}, the resulting surface area and volume equations can be formulated in the discrete versions, expressed as follows:   
	
	\begin{equation}
	S(\Sigma) =  2\pi \sum_{k=1}^{n-1} g\left( \frac{y_k + y_{k+1}}{2} \right)
	\quad\quad\quad\quad    \sqrt{1+ \left[\frac{g(y{k+1}) - g(y_{k})}{g(y_k)} \right]^2} \Delta y_k
	\end{equation}
	
	\noindent and 
	
	\begin{equation}
	\displaystyle{V(\Sigma) = 2\pi \Sigma_{k=1}^{n-1} \left[g \left(\frac{y_{k} + y_{k+1}}{2} \right) \right]^2 \Delta y_{k}}.
	\end{equation}
	
	Lastly, the final results are obtained by averaging all the geometric quantities of all the frames in each video. 
	
	\subsection{The long axis perpendicular to the $y$-axis (horizontal case,  Figure \ref{fig:xyz} (b).)}
\begin{figure}[t!]
		\centering
		\includegraphics[width=1\linewidth]{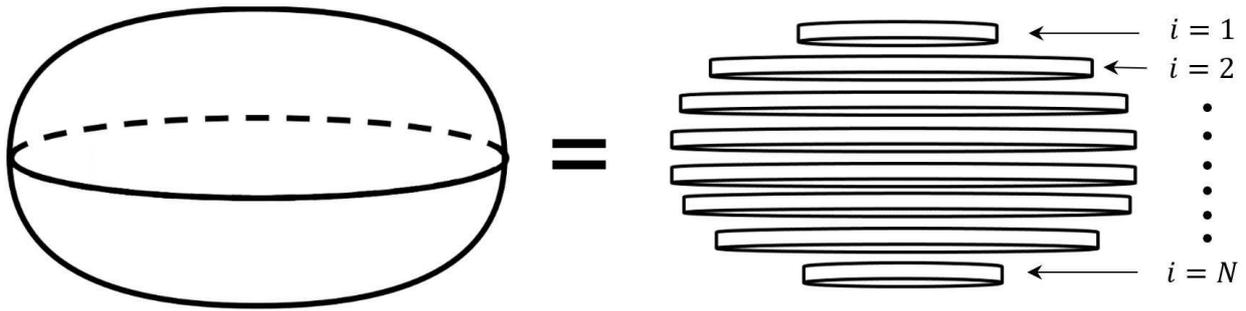}
		\caption{Schematic representations of the slices of a horizontal ham. }
		\label{fig:sumvolume}
	\end{figure}

\begin{figure}[t!]
		\centering
		\includegraphics[width=1\linewidth]{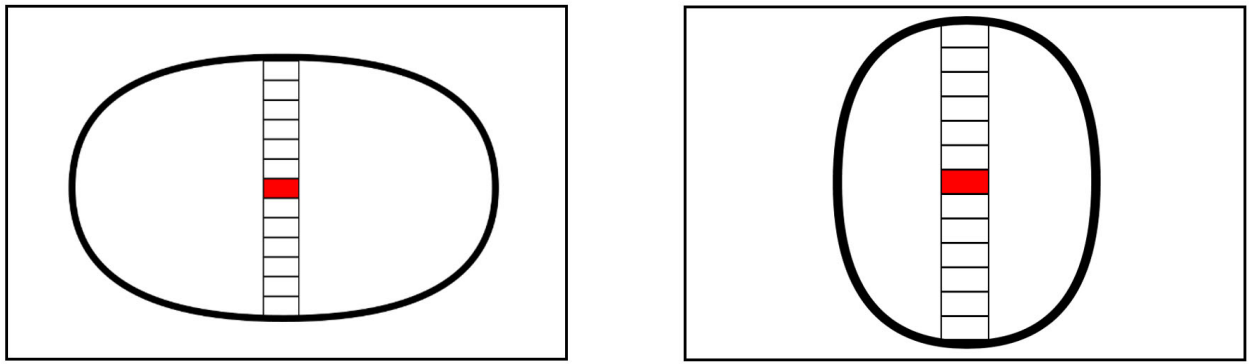}
		\caption{Illustration of the central-and-vertical lines and the centers of the ham at different frames in the video.} 
		\label{fig:center}
	\end{figure}

\begin{figure}[t!]
		\centering
		\includegraphics[width=1\linewidth]{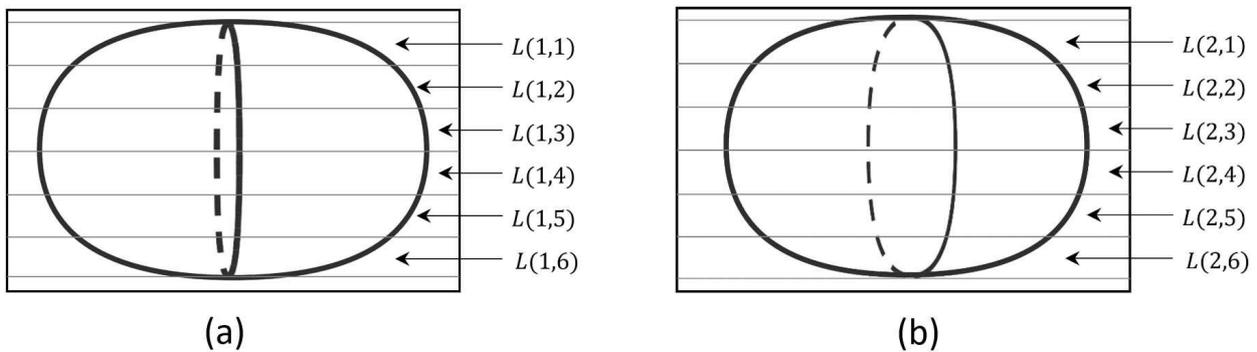}
		\caption{Two example frames selected from a video sequence where the ham is rotated at a constant angle.}
		\label{fig:2frames}
	\end{figure}

	This subsection explains the method of calculating the volume of the ham by horizontally placing it on a turntable. 
	When the table is rotating, there are a series of projected boundaries captured by the camera.
	Note that calculating the volume of ham by using the method of solid of revolution may not be applicable in our case.
	Here is the way to calculate the volume of ham in this setting and is succinctly summarized in the following three steps:
	
	\begin{enumerate}
		\item The volume of the ham can be calculated by summing the volumes of the horizontal slicing pieces from the ham (as shown in Figure \ref{fig:sumvolume}).

		\item In Figure \ref{fig:sumvolume}, it is known that the volume of each ham slice can be calculated as: 
		\begin{equation}
		\text{Volume}  = \text{Area}  \times \text{Thickness}.
		\end{equation}
		
		Note that, the smaller the thickness, the closer the predicted volume to the reference volume (a.k.a. actual value) and thus the more accurate the volume prediction.
		
		\item To obtain the area of each slice, the steps are listed as follows:
		\begin{enumerate}
			\item Recall that the images of ham is recorded with a camera. 
			In other words, it is made up of temporal sequence of discrete frames.
			For each single frame, we do not assume the center of the ham boundary is fixed at the origin and thus an adequate calibration is required to be carried out.
			The calibration steps are described as follows:
			\begin{enumerate}
				\item In each frame, the boundary of the ham as well as the central-and-vertical line of the ham are obtained by using the method as similar as described in the previous section namely, the vertical $l$ as in \eqref{eq:l}. The center of the ham in the $i$-th frame can be obtained by
				\begin{equation}
				C_i(l) = \frac{(max(l) + min(l))}{2},
				\end{equation}
				\noindent where $max(l)$ and $min(l)$ is the highest and lowest points on the central-and-vertical line $l$. Also denote $H=H(i)$ by the length of the central-and-vertical line in the $i$-th frame.
				~
				Once the line is discovered, the center of the line can be derived and set as the center of the ham. 
				The schematic diagram is drawn in Figure \ref{fig:center} to better visualize the central-and-vertical line of ham. The row consists of rectangles is the central-and-vertical line and the red one is the center of the ham. 
				\item Once the central-and-vertical lines of all the frames are determined, the next step is to identify the shortest, the average, and the longest length of the lines. We observe that when the ham is rotating, the distances of the ham to the camera are also changed.
				Specifically, the closer to the camera, the more the pixel values occupied by the ham object (the size of the ham seems to be larger). 
				Since the volume is calculated by summing the number of the pixel values, the step of appropriate calibration is needed to maintain the dynamical consistency with a given video.
				Thus, the sizes of all frames are adjusted such that they are following the length of the lines in the video same as the smallest, the average, and the longest one. Hence, there are three sets $\mathcal{S}_{short}$, $\mathcal{S}_{avg}$, $\mathcal{S}_{long}$ of adjusted frames and all frames in $\mathcal{S}_{short}$ own the same length of central-and-vertial lines which all are equal to the constant $H_{short}$; same as the lengths in $\mathcal{S}_{avg}$ and $\mathcal{S}_{long}$ are constants $H_{avg}$ and $H_{long}$ respectively. 
			\end{enumerate}
			\item The horizontal length of the $j$-th slice in the $i$-th frame is denoted as $L(i,j)$, as depicted in Figure \ref{fig:2frames}, where $i$ ranges from $1$ to $N$, the total number of frames of the video, and $j=1,\cdots, H$ is regarded as the parameter that affects the accuracy for estimating the volume. 
			\item The area of each ham slice can be derived by utilizing the formula from Integral Geometry (\cite{santalol}, Equation (1.8) in page 4):   
			\begin{equation} \label{eqn4_5}
			A(j) = \frac{1}{4} \int_{0}^{\pi} \left[L(s, j) \right]^2 + \left[\frac{\partial L(s, j)}{\partial s} \right]^2ds.
			\end{equation}
			\noindent The thickness of each ham slice determining the number of partitions on the central-and-vertical line. Note that we choose the equal thickness for all slices in all frames.
		\end{enumerate}
		
		\item Finally, the volume of the ham is molded to the following equation:
		\begin{equation}
		\text{Volume} = \sum_{j=1}^{H} \Big( A(j)\cdot \text{ thickness } \Big).
		\end{equation}
		\noindent At the current stage of derivation, three variants of volume for each ham are acquired.
		Concisely, the volume values are depending on the choices of the calibrations of ``shortest", ``average" or ``longest".
		
	\end{enumerate}

	\section{Results and discussion} 
	\label{sec:result}
	
	In this section, we present the computed volume results with detailed analysis and benchmarking against reference values.
	In addition, the procedure to obtain the benchmark values is explained in Section~\ref{subsec: water}. 
	
	\subsection{Manual Measurement of Volumes}
	\label{subsec: water}
	The volume of the hams can be measured manually using the Archimedes' principle, which is a traditional water displacement method. 
	It is a convenient solution in determining the volume especially for an irregularly shaped object.
	There are four steps to experimentally measure the volume of the hams: 
	\begin{enumerate}
		\item Partially fill the graduated container with water and record the water level.
		\item Gently immerse the ham into the water and record the water level.
		\item Calculate the volume of the ham by obtaining the difference of the two water levels.
		\item Repeat step (1) to (3) for each sample.
	\end{enumerate}
	The four steps above are repeated for twice for each of the hams.
	Then, the mean of the two measured values is computed as the final volume value.
	However, it is claimed by Wang et al.~\cite{wang2007low} that this water displacement method is not providing the perfect volume value.
	This is because it may not suitable for the objects that can absorb water, such as bread. 
	Nevertheless, since all the hams in the experiment are water resistant, such type of manual measurement is good enough as the baseline reference indicator in this experiment.
	
	\subsection{Result and Discussion}
	In order to evaluate the performance of the designed models, a comparison table against the manual measurement is reported in Table~\ref{table:result}.
	Concisely, three volume measurements of each ham are presented, namely:
	1) manual measurement;
	2) long axis parallel to the $y$-axis (vertical case), and;
	3) long axis perpendicular to the $y$-axis (horizontal case).
	Succinctly, the volumes of the vertical hams are computed using the derived method, whereas, the horizontal hams are calculated with the partitioned method.
	Inspired by the volume prediction performance reported by Du and Sun~\cite{du2006estimating}, we attempt to estimate the volumes of the hams by assuming them to be ellipsoid-shaped. 
	Unfortunately, it appears that the assumption to the above statement to be negative.
	Mathematically, it is well-known that the volume of an ellipsoid can be expressed as:
	
	\begin{equation}\label{volumeellip}
	V=\frac{\pi}{6}abc,
	\end{equation}
	
	\noindent where $a,b,c$ are the length of the axes of the ellipsoid. By using the computer vision system developed by ourselves and \eqref{volumeellip}, the axes of eight hams are measured and the detected volumes are far away from the manual. We also conclude that the shapes of all hams differ from being ellipsoidal. 
	
	It can be seen that in Table~\ref{table:result}, the mean volume predicted error of the vertical ham is 9.5\%, while the mean error of the horizontal ham is 30.26\%.  
	Reported from ~\cite{du2006estimating}, the means of error (ME) for the partitioned method and derived method are given by -1.79\% and -4.96\%, respectively.
	However, there are some differences in the calculation method and the experimental setup between their work with this study. 
	Firstly, the value of ME that reported by ~\cite{du2006estimating} is the average error, which is not equivalent to the absolute means error.
	The formula of the mean absolute error (MAE) employed in this study is written generically as:
	
	\begin{equation}
	\label{eq:mae}
	\text{MAE} = \frac{1}{n}\sum\limits_{i=1}^{n}|e_i|
	\end{equation}
	
	\noindent  where $n$ is the total number of ham, $e_i$ is the volume difference of the $i$-th ham between the manual measurement and the predicted value. The smaller MAE implies that the smaller ME. However, the reversed is not true. 
	The reason of using MAE in this study is to calculate how close the average estimated volume to the actual value.  
	
	Secondly, the range of the hams used in~\cite{du2006estimating} is from approximately 920$cm^3$ to 1060$cm^3$, whereas the minimum volume of the ham in this experiment is 70$cm^3$ and the maximum volume is 500$cm^3$. In this study, we are also concerned about the distribution of the hams' volumes. The standard deviation of the sample mean distribution are calculated by using the formula given in Equation \eqref{eq:SE} in order to investigate the distribution of the volumes of the hams.
	
	\begin{equation}
	\label{eq:SE}
	{\text SE} = \frac{\sigma }{\sqrt{n}}.
	\end{equation}
	
	\noindent From the comparison, it has a very large difference of standard deviation for the volumes of the sample sizes in both the experiments.
	Concretely, the standard deviation volume in~\cite{du2006estimating} is about 47$cm^3$, while in this experiment is approximately 184$cm^3$.
	It is reasonable that similar shapes and sizes of the ham samples tend to generate better prediction accuracies.
	
	Thirdly, the proposed method employed a deep learning architecture which is always applied on big data experiments. 
	In contrary,~\cite{du2006estimating} adopted handcrafted features which the configuration settings of the mechanism proposed can be heuristically fine-tuned to best fit the train and test samples.
	Therefore, it is not surprised that~\cite{du2006estimating} achieve better performance compared to ours.
	With only eight sample of hams, our method is able to exhibit reasonable good performance and it is believe that it will lead to significantly better results if the sample size increases.
	Furthermore, deep learning model is famed with good generalization ability and hence the proposed method may be applicable to measure volumes of hams with different shapes and sizes.
	
\begin{figure}[h]
		\centering
		\includegraphics[width=1\linewidth]{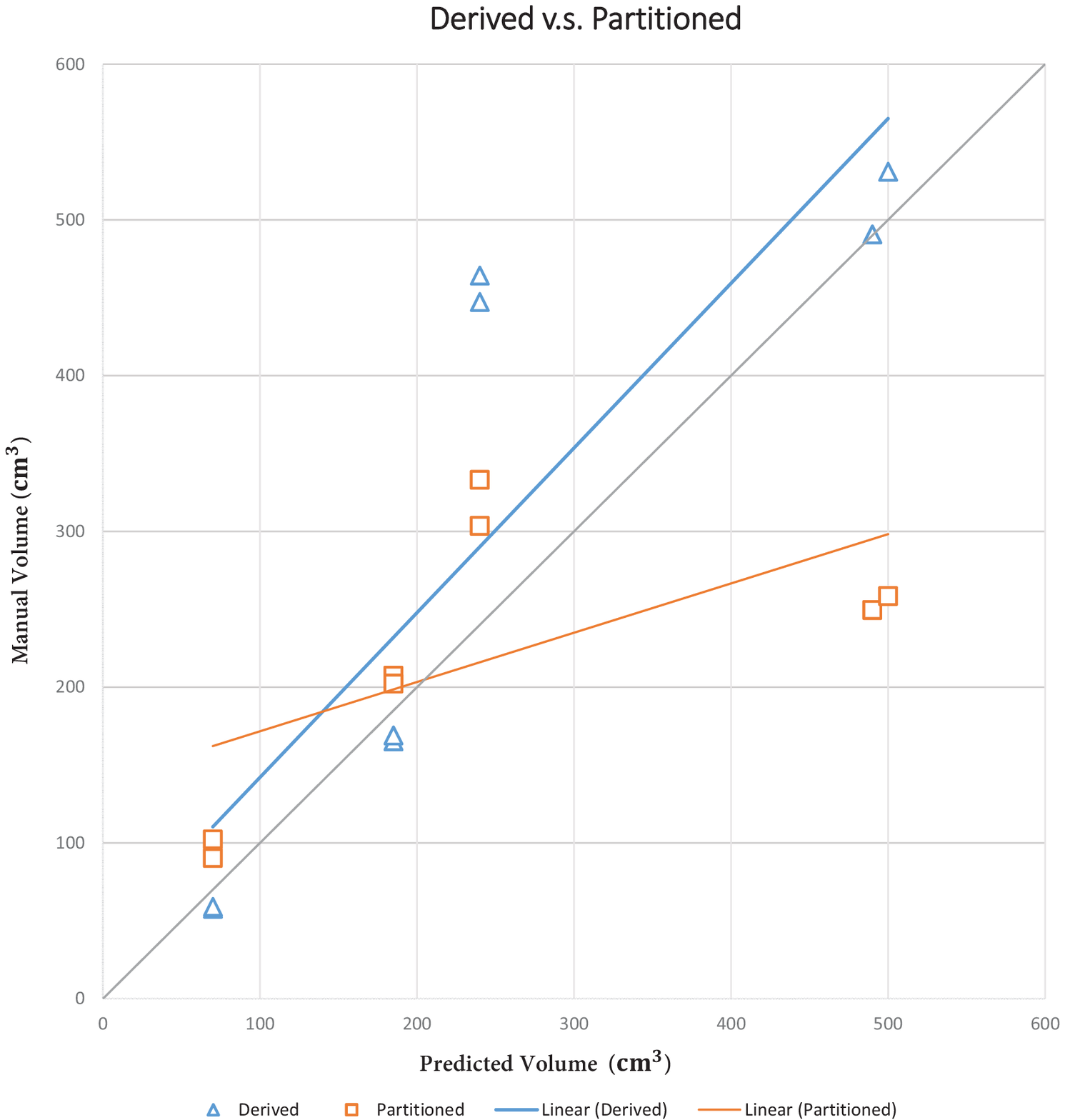}
		\caption{Performances of the developed two methods for predicting the volume of ham}
		\label{fig:chart1}
	\end{figure}

	It is observed in Table~\ref{table:result} that the average volume prediction performance of the vertical ham is better than the horizontal. 
	Specifically for ham \# 1, the volume prediction accuracy for the horizontal position is the highest (i.e., 99.86\%).
	However, it appears to be the worst result (i.e., accuracy of 50.93\%) when it is evaluated horizontally using the partitioned method.
	This is because the ham's shapes do not change much as it is vertically rotating in $\degree{360}$, as shown in Figure~ \ref{fig:2_1} 
	(from frame 10 to frame 240).
	On the contrary, the shapes for the horizontal ham in the video are changed dramatically between the frame 10 to frame 240 when it is being captured at the fixed rotating stand.
	It is obvious that the shapes of the ham are totally different for frame 40 and frame 160 in Figure~\ref{fig:2_1}, which shows the circle and the ellipse respectively.
	To further analyze the shape of the ham in each frame, Table \ref{table:dimensionaxes} shows the lengths of the axes of all samples measured by the computer vision system (the details are explained in Section \ref{sec:CVS}), also tabulated in Table~\ref{table:dimensionaxes}.
	Among eight samples, the ratio of three axes (i.e., the longest, shorter 1 and shorter 2) is far away from $1:1:1$, which means that when horizontally rotating the ham the shapes of the boundaries indeed affect the accuracy of our prediction. 
	It is observed that the computed results are not similar to the manual measurements.
	In order to provide a better understanding of the overall performance of the proposed method, Figure \ref{fig:chart1} 
	shows predicted volume by the proposed method versus the actual volume. 
	It can be observed that the performance of the derived method is better that of partitioned method.
	
	\begin{table*}[tb]
		\begin{center}
			\footnotesize
			\caption{Comparison of the volume prediction with the manual, derived and partitioned methods.}
			\centering
			\label{table:result}
			\begin{tabular}{cccccccccc}
				\noalign{\smallskip}
				\hline
				\noalign{\smallskip}

				\multirow{2}{*}{Ham} 
				& Manual
				&
				& \multicolumn{2}{c}{Derived (Vertical)} 
				&
				& \multicolumn{2}{c}{Partitioned (Horizontal)} 
				
				\\
				
				\cmidrule{2-2} \cmidrule{4-5}\cmidrule{7-8}
				\noalign{\smallskip}
				
				& Volume ($cm^3$)
				&
				& Volume ($cm^3$)
				& Error (\%)
				&
				& Volume ($cm^3$)
				& Error (\%)
				\\
				
				\noalign{\smallskip}
				\hline
				\noalign{\smallskip}
				1  
				&  490 
				&
				& 490.69  
				& 0.14 
				&
				&  249.51
				&  49.07 
				\\
				
				2 
				& 500 
				&
				& 530.94  
				& 6.19 
				&
				& 258.45 
				& 48.31 
				\\
				
				3  
				& 420 
				&
				& 464.27  
				& 10.54
				&
				& 303.50 
				& 27.74 
				\\
				
				4 
				&  420 
				& 
				&  447.31 
				& 6.50
				&
				& 333.10 
				& 20.48 
				\\
				
				5  
				&  185  
				&
				&  165.26 
				& 10.67
				&
				& 207.31 
				& 12.06 
				\\ 
				
				6
				& 185  
				&
				& 169.04
				& 8.65 
				&
				& 202.28 
				& 9.34 
				\\
				
				7 
				& 70 
				&
				& 57.74 
				& 17.51 
				&
				&  90.33 
				& 29.04 
				\\
				
				8 
				& 70  
				&
				& 58.95 
				& 15.79 
				&
				& 102.23 
				& 46.04 
				\\
				
				\noalign{\smallskip}
				\hline
				\noalign{\smallskip}
				
				Mean
				& 292.50 
				&
				& 298.03 
				& 9.50 
				&
				& 218.34 
				& 30.26 %
				\\
				
				Standard Deviation
				& 172.01
				& 
				& 190.54    
				&  5.18          
				&
				& 81.59            
				& 15.00           
				\\
				\hline
				
			\end{tabular}
			
		\end{center}
	\end{table*}

	\begin{table*}[h]
		\begin{center}
			\caption{Lengths of axes for the hams measured by the computer vision system}
			\label{table:dimensionaxes}
			\begin{tabular}{cccccc}
				\noalign{\smallskip}
				\hline
				\noalign{\smallskip}
				\multirow{2}{*}{Ham} 
				& Longest Axes 
				& Shorter Axes 1
				& Shorter Axes 2
				& Volume, $V$
				& \multirow{2}{*}{\% of manual} \\

				& ($cm$)
				& ($cm$)
				& ($cm$)
				& ($cm^3$)
				& \\
				
				\noalign{\smallskip}
				\hline
				\noalign{\smallskip}
				
				1
				&717.0
				&459
				&457
				&246.3
				&53.3\\
				
				\noalign{\smallskip}
				2
				&736.5
				&463
				&467
				&260.8
				&52.2\\
				
				\noalign{\smallskip}
				3
				&881.5
				&383
				&390
				&215.7
				&51.4\\

				\noalign{\smallskip}
				4
				&877.0
				&385
				&381
				&210.7
				&50.2\\
				
				\noalign{\smallskip}
				5
				&774.0
				&251
				&240
				&76.5
				&41.3\\
				
				\noalign{\smallskip}
				6
				&810.5
				&241
				&235
				&75.2
				&40.6\\
				
				\noalign{\smallskip}
				7
				&732.5
				&139
				&141
				&23.5
				&33.6\\
				
				\noalign{\smallskip}
				8
				&725.5
				&151
				&142
				&25.5
				&36.5\\
				
				\hline
				
			\end{tabular}
		\end{center}
	\end{table*}

	\section{Conclusion} 
	\label{sec:conclusion}
	
	In this paper, we demonstrate that the values of the geometric properties (i.e., the boundaries, centers, volume, etc.) can be effectively acquired from the image processing techniques with deep learning.
	Succinctly, a series of methods that involve the computer vision and the mathematical modeling is introduced to perform the volume prediction of the hams.
	Without the need of physical touching the object and the prior knowledge of the object's shape or size, its volume can be obtained by simply having the two-dimensional photos of that object at different angles.
	The algorithm proposed is modified from a popular instance segmentation approach, namely Mask-RCNN.
	We evaluated the proposed method on four types of hams with different length, height and thickness that further highlights the robustness of the proposed method.
	Particularly, we present the volume prediction performance in two observatory viewpoints of the hams: vertically and horizontally.
	The results indicated that the overall performance of vertical viewpoint yields better than the horizontal ones.
	Specifically, the best volume prediction accuracy attained for the vertical viewpoint of up to 99.86\%, whereas 90.66\% for the horizontal viewpoint.
	To summarize, the mean absolute error of the vertically placed ham is 9.5\% while the horizontal ham is 30.26\%.
	Nevertheless, this is first work to test on ham object with a variety of object's shape with small sample size.
	In addition, we adopt deep learning method to intuitive learn the object's features from the input image data.

	For future works, attentions can be devoted to the following possible directions which potentially lead to accuracy improvement is volume prediction task.
	Firstly, to calibrate of distances between the hams and the camera when capturing the videos.
	This is because there may exist human error where the ham might not be placed exactly at the same location on the rotating display stand every time.
	Secondly, the computation of the surface area of the ham is possible by adopting an analogous method.
	Thirdly, the machine learning training model provides flexible and adaptive modeling approaches that could improve the prediction accuracy.
	The experiments can be extended to test on other object such as in the variation of object's features such as color, texture or material composition.
	In such situations, increasing the sample size is expecting to facilitate the object's learning progress and enhance the discriminant ability of the training model.
	Last but not least, it is also interesting to study the optimization methods and difficulties involved in deep learning to achieve better generalization capabilities.
	
	\section{Acknowledgement} 
	This work was funded by Ministry of Science and Technology (MOST) (Grant Number: MOST 107-2218-E-035-006-) and also by Xinyang Normal University.
	The authors would also like to thank Hsiu-Chi Chang, Yu-Siang Su and Jih-Hsiang Cheng for their assistance in the sample preparation, experimental setup  and image acquisition.

	
	\bibliographystyle{elsarticle-num}
	\bibliography{mybibfile}

\begin{thebibliography}{10}
\expandafter\ifx\csname url\endcsname\relax
\def\url#1{\texttt{#1}}\fi
\expandafter\ifx\csname urlprefix\endcsname\relax\def\urlprefix{URL }\fi
\expandafter\ifx\csname href\endcsname\relax
\def\href#1#2{#2} \def\path#1{#1}\fi

\bibitem{prasad2009determinants}
C.~J. Prasad, A.~Aryasri, Determinants of shopper behaviour in e-tailing: An
empirical analysis, Paradigm 13~(1) (2009) 73--83.

\bibitem{ups2017}
Ups pulse of the online shopper, executive summary, european study, (2017)\\
\url{https://www.ups.com/assets/resources/media/en_gb/ups_potos_eu_en_linked.pdf}

\bibitem{mccabe2003effect}
D.~B. McCabe, S.~M. Nowlis, The effect of examining actual products or product
descriptions on consumer preference, Journal of Consumer Psychology 13~(4)
(2003) 431--439.

\bibitem{hoyer1984examination}
W.~D. Hoyer, An examination of consumer decision making for a common repeat
purchase product, Journal of consumer research 11~(3) (1984) 822--829.

\bibitem{ronneberger2015u}
O.~Ronneberger, P.~Fischer, T.~Brox, U-net: Convolutional networks for
biomedical image segmentation, in: International Conference on Medical image
computing and computer-assisted intervention, Springer, 2015, pp. 234--241.

\bibitem{liong2018less}
S.-T. Liong, J.~See, K.~Wong, R.~C.-W. Phan, Less is more: Micro-expression
recognition from video using apex frame, Signal Processing: Image
Communication 62 (2018) 82--92.

\bibitem{pfister2014comparative}
A.~Pfister, A.~M. West, S.~Bronner, J.~A. Noah, Comparative abilities of
microsoft kinect and vicon 3d motion capture for gait analysis, Journal of
medical engineering \& technology 38~(5) (2014) 274--280.

\bibitem{rao2015computer}
R.~S. Rao, S.~T. Ali, A computer vision technique to detect phishing attacks,
in: communication systems and network technologies (CSNT), 2015 fifth
international conference on, IEEE, 2015, pp. 596--601.

\bibitem{ziaratban2017modeling}
A.~Ziaratban, M.~Azadbakht, A.~Ghasemnezhad, Modeling of volume and surface
area of apple from their geometric characteristics and artificial neural
network, International Journal of Food Properties 20~(4) (2017) 762--768.

\bibitem{soltani2015egg}
M.~Soltani, M.~Omid, R.~Alimardani, Egg volume prediction using machine vision
technique based on pappus theorem and artificial neural network, Journal of
food science and technology 52~(5) (2015) 3065--3071.

\bibitem{anders2015numerical}
A.~Anders, Z.~Kaliniewicz, P.~Markowski, Numerical modelling of agricultural
products on the example of bean and yellow lupine seeds, International
Agrophysics 29~(4) (2015) 397--403.

\bibitem{lee2015design}
D.~Lee, K.~Lee, S.~Kim, Y.~Yang, Design of an optimum computer vision-based
automatic abalone (haliotis discus hannai) grading algorithm, Journal of food
science 80~(4) (2015) E729--E733.

\bibitem{du2006estimating}
C.-J. Du, D.-W. Sun, Estimating the surface area and volume of ellipsoidal ham
using computer vision, Journal of Food Engineering 73~(3) (2006) 260--268.

\bibitem{volumeoverview2015}
A.~G. . D.~G. G~Vivek~Venkatesh, S Md~Iqbal, Estimation of volume and mass of
axi-symmetric fruits using image processing technique, International Journal
of Food Properties 18.

\bibitem{rasband2011imagej}
W.~S. Rasband, Imagej, us national institutes of health, bethesda, maryland,
usa, http://imagej. nih. gov/ij/.

\bibitem{davis1975survey}
L.~S. Davis, A survey of edge detection techniques, Computer graphics and image
processing 4~(3) (1975) 248--270.

\bibitem{rosin1993note}
P.~L. Rosin, A note on the least squares fitting of ellipses, Pattern
Recognition Letters 14~(10) (1993) 799--808.

\bibitem{asadi2010egg}
V.~Asadi, M.~Raoufat, Egg weight estimation by machine vision and neural
network techniques (a case study fresh egg)., International Journal of
Natural \& Engineering Sciences 4~(2).

\bibitem{thomas2005thomas}
G.~Thomas, M.~Weir, J.~Hass, F.~Giordano, Thomas’ calculus including
second-order differential equations (2005).

\bibitem{waranusast2016egg}
R.~Waranusast, P.~Intayod, D.~Makhod, Egg size classification on android mobile
devices using image processing and machine learning, in: Student Project
Conference (ICT-ISPC), 2016 Fifth ICT International, IEEE, 2016, pp.
170--173.

\bibitem{dalal2005histograms}
N.~Dalal, B.~Triggs, Histograms of oriented gradients for human detection, in:
Computer Vision and Pattern Recognition, 2005. CVPR 2005. IEEE Computer
Society Conference on, Vol.~1, IEEE, 2005, pp. 886--893.

\bibitem{rother2004grabcut}
C.~Rother, V.~Kolmogorov, A.~Blake, Grabcut: Interactive foreground extraction
using iterated graph cuts, in: ACM transactions on graphics (TOG), Vol.~23,
ACM, 2004, pp. 309--314.

\bibitem{otsu1979threshold}
N.~Otsu, A threshold selection method from gray-level histograms, IEEE
transactions on systems, man, and cybernetics 9~(1) (1979) 62--66.

\bibitem{canny1986computational}
J.~Canny, A computational approach to edge detection, IEEE Transactions on
pattern analysis and machine intelligence~(6) (1986) 679--698.

\bibitem{de1978practical}
C.~De~Boor, C.~De~Boor, E.-U. Math{\'e}maticien, C.~De~Boor, C.~De~Boor, A
practical guide to splines, Vol.~27, Springer-Verlag New York, 1978.

\bibitem{OKT}
A.~T. M.~Omid, M.~Khojastehnazhand, Estimating volume and mass of citrus fruits
by image processing technique, Journal of Food Engineering 100~(2) (2010)
315–321.

\bibitem{anovelmethodegg}
Z.~Q.~Y. Weizheng~Zhang, Xiang~Wu, A novel method for measuring the volume and
surface area of egg, Journal of Food Engineering 170 (2016) 160--169.

\bibitem{MonteCarlo1}
A.~A. B.~I. Joko~Siswantoro, Anton Satria~Prabuwono, Monte carlo method with
heuristic adjustment for irregularly shaped food product volume measurement,
The Scientific World Journal 2014~(http://dx.doi.org/10.1155/2014/683048)
(2014) 10 pages.

\bibitem{MonteCarlo2}
A.~A. Joko~Siswantoro, Anton Satria~Prabuwono, Volume measurement of food
product with irregular shape using computer vision and monte carlo method: A
framework, Procedia Technology 11 (2013) 764--770.

\bibitem{deterorange1}
A.~T. M.~Khojastehnazhand, M.~Omid, Determination of orange volume and surface
area using image processing technique, International Agrophysics 23 (2009)
237--242.

\bibitem{schwartz1993pappus}
R.~Schwartz, Pappus’s theorem and the modular group, Publications
Math{\'e}matiques de l'Institut des Hautes {\'E}tudes Scientifiques 78~(1)
(1993) 187--206.

\bibitem{he2017mask}
K.~He, G.~Gkioxari, P.~Doll{\'a}r, R.~Girshick, Mask r-cnn, in: Computer Vision
(ICCV), 2017 IEEE International Conference on, IEEE, 2017, pp. 2980--2988.

\bibitem{ren2015faster}
S.~Ren, K.~He, R.~Girshick, J.~Sun, Faster r-cnn: Towards real-time object
detection with region proposal networks, in: Advances in neural information
processing systems, 2015, pp. 91--99.

\bibitem{girdhar2018detect}
R.~Girdhar, G.~Gkioxari, L.~Torresani, M.~Paluri, D.~Tran, Detect-and-track:
Efficient pose estimation in videos, in: Proceedings of the IEEE Conference
on Computer Vision and Pattern Recognition, 2018, pp. 350--359.

\bibitem{malhotra2018autonomous}
K.~R. Malhotra, A.~Davoudi, S.~Siegel, A.~Bihorac, P.~Rashidi, Autonomous
detection of disruptions in the intensive care unit using deep mask r-cnn,
in: Proc. IEEE Conference on Computer Vision and Pattern Recognition
Workshops, 2018, pp. 1863--1865.

\bibitem{wei2016mask}
X.-S. Wei, C.-W. Xie, J.~Wu, Mask-cnn: Localizing parts and selecting
descriptors for fine-grained image recognition, arXiv preprint
arXiv:1605.06878.

\bibitem{loudon2018detecting}
J.~S. Loudon, Detecting and localizing cell nuclei in medical images., Master's
thesis, NTNU (2018).

\bibitem{lin2017feature}
T.-Y. Lin, P.~Doll{\'a}r, R.~B. Girshick, K.~He, B.~Hariharan, S.~J. Belongie,
Feature pyramid networks for object detection., in: CVPR, Vol.~1, 2017, p.~4.

\bibitem{he2016deep}
K.~He, X.~Zhang, S.~Ren, J.~Sun, Deep residual learning for image recognition,
in: Proceedings of the IEEE conference on computer vision and pattern
recognition, 2016, pp. 770--778.

\bibitem{imagenet_cvpr09}
J.~Deng, W.~Dong, R.~Socher, L.-J. Li, K.~Li, L.~Fei-Fei, {ImageNet: A
Large-Scale Hierarchical Image Database}, in: CVPR09, 2009.

\bibitem{lin2014microsoft}
T.-Y. Lin, M.~Maire, S.~Belongie, J.~Hays, P.~Perona, D.~Ramanan,
P.~Doll{\'a}r, C.~L. Zitnick, Microsoft coco: Common objects in context, in:
European conference on computer vision, Springer, 2014, pp. 740--755.

\bibitem{matterport_labelme_2016}
K.~Wada, Image polygonal annotation with python (polygon, rectangle, circle,
line, point and image-level flag annotation),
\url{https://github.com/wkentaro/labelme} (2016).

\bibitem{matterport_maskrcnn_2017}
W.~Abdulla, Mask r-cnn for object detection and instance segmentation on keras
and tensorflow, \url{https://github.com/matterport/Mask_RCNN} (2017).

\bibitem{kresta2015advances}
S.~M. Kresta, A.~W. Etchells~III, V.~A. Atiemo-Obeng, D.~S. Dickey, Advances in
industrial mixing: a companion to the handbook of industrial mixing, John
Wiley \& Sons, 2015.

\bibitem{santalol}
L.~Santal{\'o}, Integral Geometry and Geometric Probability, Vol. (Encyclopedia
of mathematics and its applications; v. 1, SAddison-Wesley Publishing
Company, 1976.

\bibitem{wang2007low}
T.~Y. Wang, S.~K. Nguang, Low cost sensor for volume and surface area
computation of axi-symmetric agricultural products, Journal of Food
Engineering 79~(3) (2007) 870--877.

\end{thebibliography}
	
	
	
	
	
	
	
\end{document}